\documentclass[runningheads]{llncs}
\usepackage{graphicx}

\usepackage{tikz}
\usepackage{comment}
\usepackage{amsmath,amssymb} 
\usepackage{color}

\usepackage{booktabs}
\usepackage{multirow}
\usepackage{array}
\usepackage{bm}
\usepackage{diagbox}
\usepackage[capitalize]{cleveref}
\usepackage[titletoc]{appendix}

\crefname{section}{Sec.}{Secs.}
\Crefname{section}{Section}{Sections}
\Crefname{table}{Table}{Tables}
\crefname{table}{Table.}{Tabs.}

\usepackage{float}
\newfloat{figtab}{htb}{fgtb}
\makeatletter
  \newcommand\figcaption{\def\@captype{figure}\caption}
  \newcommand\tabcaption{\def\@captype{table}\caption}
\makeatother

\usepackage[accsupp]{axessibility}  

\makeatletter
\def\@fnsymbol#1{\ensuremath{\ifcase#1\or * \or \dagger\or \ddagger\or
   \mathsection\or \mathparagraph\or \|\or **\or \dagger\dagger
   \or \ddagger\ddagger \else\@ctrerr\fi}}
 \makeatother

\begin{document}
\pagestyle{headings}
\mainmatter
\def\ECCVSubNumber{6275}  

\authorrunning{ }
\titlerunning{ }
\title{A Simple Baseline for Open-Vocabulary Semantic Segmentation with \\Pre-trained Vision-language Model} 
\author{
Mengde~Xu\inst{1,3}\thanks{~Equal contribution.} \and 
Zheng~Zhang\inst{1,3}$^\ast$ \and 
Fangyun~Wei\inst{3}$^\ast$ \and 
Yutong~Lin\inst{2,3} \and \\
Yue~Cao\inst{3} \and
Han~Hu\inst{3} \and
Xiang~Bai\inst{1}\thanks{~Corresponding author}
}
\institute{Huazhong University of Science and Technology \and Xi’an Jiaotong University \and Microsoft Research Asia}



\maketitle

\begin{abstract}
Recently, open-vocabulary image classification by vision language pre-training has demonstrated incredible achievements, that the model can classify arbitrary categories without seeing additional annotated images of that category. However, it is still unclear how to make the open-vocabulary recognition work well on broader vision problems. This paper targets open-vocabulary semantic segmentation by building it on an off-the-shelf pre-trained vision-language model, i.e., CLIP. However, semantic segmentation and the CLIP model perform on different visual granularity, that semantic segmentation processes on pixels while CLIP performs on images. To remedy the discrepancy in processing granularity, we refuse the use of the prevalent one-stage FCN based framework, and advocate a two-stage semantic segmentation framework, with the first stage extracting generalizable mask proposals and the second stage leveraging an image based CLIP model to perform open-vocabulary classification on the masked image crops which are generated in the first stage. Our experimental results show that this two-stage framework can achieve superior performance than FCN when trained only on COCO Stuff dataset and evaluated on other datasets without fine-tuning. Moreover, this simple framework also surpasses previous state-of-the-arts of zero-shot semantic segmentation by a large margin: +29.5 hIoU on the Pascal VOC 2012 dataset, and +8.9 hIoU on the COCO Stuff dataset. With its simplicity and strong performance, we hope this framework to serve as a baseline to facilitate future research. The code are made publicly available at~\url{https://github.com/MendelXu/zsseg.baseline}. 
\end{abstract}

\section{Introduction}
Semantic segmentation is a fundamental computer vision task that assigns every pixel of an image with category labels. Accompanied by the development of deep learning~\cite{NIPS2012_c399862d,simonyan2014very,he2016deep,dosovitskiy2020image,liu2021swin}, the semantic segmentation has also evolved tremendously under the supervised learning paradigm~\cite{long2015fully,chen2017deeplab,badrinarayanan2017segnet}. However, unlike common image-level datasets such as ImageNet-1K/ImageNet-22K image classification which are easily scaled up to tens of thousands of categories, existing semantic segmentation tasks involve usually up to tens or hundreds of categories due to the significantly higher annotation cost, and thus limit the segmentors' capability in handling rich semantics. 

Zero-shot semantic segmentation~\cite{bucher2019zero} is an attempt to break the bottleneck of limited categories. However, the narrowly defined zero-shot semantic segmentation usually only takes a small amount of labeled segmentation data and refuses to make use of any other data/information, consequently resulting in poor performance. In this work, we focus on another more practical setting: \emph{open-vocabulary} semantic segmentation, as a generalized zero-shot semantic segmentation, concentrates more on establishing a feasible method to segment arbitrary classes and allows the use of additional data/information except the segmentation data. Specifically, we propose to leverage a recent advance of image-level vision-language learning model, i.e., CLIP~\cite{radford2021learning}.

While the vision-language learning model has learnt a strong vision-category alignment model using rich image-caption data, how to effectively transfer its image-level recognition capability to pixel-level is unclear. An natural idea is to integrate the vision-language model with a fully convolutional networks (FCN)~\cite{long2015fully}, an architecture widely used for fully supervised semantic segmentation. A main difficulty of the integration is that the CLIP model is learnt at image-level, which differs from the granularity of FCN that models semantic segmentation as a pixel classification problem, where a linear classifier is applied on each pixel feature to produce the classification results, with each column of the linear classifier weight matrix representing each category. Empirically, we found the granularity inconsistency lead unsatisfactory performance.

To better leverage the strong vision-category correspondence capability involved in the image-level CLIP model, we pursue mask proposal based semantic segmentation approaches such as MaskFormer~\cite{cheng2021per}, which first extracts a set of class-agnostic mask proposals and then classifies each mask proposal into a different category. This two-stage approach decouples the semantic segmentation task into two sub-tasks of class-agnostic mask generation and mask category classification. Both sub-tasks prove well adaptation to handle unseen classes: firstly, the class-agnostic mask proposal generation trained using \emph{seen} classes is observed well generalizable to \emph{unseen} classes; secondly, the second mask proposal classification stage is at a same recognition granularity than that used in a CLIP model. To further bridge the gap with a CLIP model, the masked image crop of each proposal is used as input to the CLIP model for \emph{unseen} classes classification. In addition, we employ a prompt-learning approach~\cite{liu2021pre} to further improve the \emph{unseen} classes classification accuracy given a pre-trained CLIP model.

We evaluate the proposed approach under two different settings: 1)\emph{Cross-dataset} setting where the model is trained on one dataset and evaluated on other datasets without fine-tuning. Under this setting, our two-stage framework demonstrate well generalization capability. It outperforms FCN approach by \textbf{+13.1} mIoU on Cityscapes, \textbf{+19.6} mIoU on Pascal Context, \textbf{+5.6} mIoU on ADE20k with 150 classes and \textbf{+2.9} mIoU on ADE20k with 847 classes. 2) \emph{Zero-shot} setting where the model is trained on a part of \emph{seen} class of a dataset and evaluated on all classes (including \emph{seen} and \emph{unseen} classes). We use this setting for comparing with other zero-shot semantic segmentation methods. We show that the proposed approach, though simple and straightforward, can surpass previous state-of-the-arts zero-shot segmentation approaches~\cite{bucher2019zero,pastore2021closer,xian2019semantic,gu2020context} by a large margin. On Pascal VOC 2012~\cite{everingham2011pascal}, this approach outperforms previous best methods that w/o self-training by \textbf{+37.8} hIoU , and by \textbf{+29.5} hIoU when an additional self-training process is involved. On COCO Stuff~\cite{caesar2018coco}, the approach outperforms previous best methods that w/o self-training by \textbf{+19.6} hIoU and by \textbf{+8.9} hIoU when an additional self-training process is involved. We hope our simple but effective approach can encourage more study in this direction.

\section{Related Works}
\subsubsection{Vision-Language Pre-training.}
Vision-language pre-training focuses on how to connect visual concepts and language concepts. Early approaches~\cite{su2019vl,lu2019vilbert,chen2019uniter,li2020unicoder,li2020oscar} were performed on some cleaned datasets with relatively small data scale. Therefore, those models usually need to be fine-tuned on some specific downstream tasks. Some recent works~\cite{radford2021learning,jia2021scaling} have explored the benefits of large-scale noisy data obtained from web pages for vision-language pre-training. CLIP~\cite{radford2021learning}, as a representative work, employs a contrastive learning approach to distinguish the correct image-text pair in each training batch. Because many vision/language concepts are covered in large-scale data, the CLIP illustrates surprisingly strong capability on zero-shot/open-vocabulary image classification and image-text retrieval. This work introduces the CLIP model as a strong vision-category correspondant for open-vocabulary semantic segmentation.

\subsubsection{Semantic Segmentation.} 
Semantic segmentation is a fundamental task in computer vision that aims to assign a category to each pixel. Fully convolutional network~\cite{long2015fully} and its variants~\cite{chen2017deeplab,yin2020disentangled,badrinarayanan2017segnet}, as a practical and straightforward approach to model the semantic segmentation as a pixel-wise classification problem, have dominated this field in the past few years. Recently, MaskFormer~\cite{cheng2021per} explored to model the semantic segmentation as two sub-tasks: segment generation and segment classification and has shown competitive performance compared to FCN based approaches. 

\subsubsection{Zero-Shot Learning and Open-vocabulary Learning.}
Zero-shot learning has been widely studied in recent years. A narrowly defined zero-shot learning focuses on learning transferable representations from the annotated data of seen classes to represent unseen classes. For example,~\cite{akata2015label,xian2016latent} proposed to learn a joint embedding space between the images and the name/description of the category for image classification, and~\cite{lampert2013attribute} explored taking the advantages of mid-level semantic representation. Recently, the open-vocabulary learning has attracted more attentions. As a generalized zero-shot learning, the open-vocabulary learning is more concerned with establishing a feasible method for arbitrary class recognition and allows the use of any additional information. For example, Visual N-Grams~\cite{li2017learning} and CLIP~\cite{radford2021learning} explored the use of web-crawled data for image classification and \cite{gu2021zero} introduced the vision-language pre-training model for the open-vocabulary object detection and showed it could significantly improve the long-tile object detection~\cite{gupta2019lvis}.

\subsubsection{Zero-shot Semantic Segmentation.}
Some pioneer works to study the zero-shot learning for semantic segmentation. ZS3Net~\cite{bucher2019zero} uses generative models to synthesize pixel-level features by word embeddings of \emph{unseen} classes. CSRL~\cite{li2020consistent} further incorporating the structural relation in feature synthesize.  CaGNet~\cite{gu2020context,gu2020pixel} introduce a contextual module for better feature generation. 
Different from ~\cite{bucher2019zero,li2020consistent,gu2020context}, SPNet~\cite{xian2019semantic} attempt to mapping vision feature to the semantic space via word embedding. JoEm~\cite{baek2021exploiting} a joint embedding strategy between the vision encoder and semantic encoder. In~\cite{kato2019zero}, variational mapping is used to learn semantic features. In~\cite{hu2020uncertainty}, the uncertainty-aware losses are proposed to eliminate noisy samples. Other works explored other directions or aspects of zero-shot semantic segmentation. In~\cite{sp2net}, the super-pixel pooling is utilized to improve the region grouping generalization. In~\cite{pastore2021closer}, the self-training for zero-shot semantic segmentation are carefully studied. In~\cite{song2018transductive,lv2020learning}, the transductive learning setting are explored. In~\cite{tian2020cap2seg} , they explore the utilization of image caption. However, all those methods have not explored the utilization of the vision-language pre-training model in zero-shot semantic segmentation. There are two concurrent work~\cite{li2022languagedriven,ghiasi2021open} try to utilize the vision-language pre-training model in semantic segmentation. However, LSeg~\cite{li2022languagedriven} is an FCN-based
approach focus on few shot setting. Openseg~\cite{ghiasi2021open}, which is a similar work to ours, utilizes external grounding dataset while we don’t. In addition, Openseg is based on ALIGN~\cite{jia2021scaling} while we adopt CLIP~\cite{radford2021learning}.

\section{Preliminary}
In this section, we first introduce the setting of open-vocabulary semantic segmentation and revisit CLIP as preliminary.

\subsection{Open-vocabulary Semantic Segmentation}
\subsubsection{Zero-Shot Setting.} Open-vocabulary is an generalized zero-shot task, so the zero-shot semantic segmentation protocol can also evaluate open-vocabulary semantic segmentation. In this setting, model predicts masks for \emph{unseen} classes ${\cal C}^{\text{unseen}}$ by learning from some labeled data of \emph{seen} classes ${\cal C}^{\text{seen}}$, and the \emph{seen} classes and \emph{unseen} classes are disjoint, i.e., ${\cal C}^{\text{unseen}}\cap{\cal C}^{\text{seen}}=\varnothing$. Usually, ${\cal C}^{\text{seen}}$ and ${\cal C}^{\text{unseen}}$ are often represented with semantic words like \textit{dog, cat, apple}, and sometimes the description of the classes are also provided. 

During training, a training set ${\cal X}_{\mathrm{train}}=\left\{ ({\cal I}_{k},{\cal M}_{k})\right\}$ with input images ${\cal I}_{k}$ and the ground-truth semantic segmentation annotations ${\cal M}_{k}$ is provided, and the training annotations \{${\cal M}_{k}$\} contains only the \emph{seen} classes. The trained model is evaluated on a testing set ${\cal X}_{\mathrm{test}}$, both \emph{seen} classes and \emph{unseen} classes need to be predicted in testing set ${\cal X}_{\mathrm{test}}$.

\subsubsection{Cross-Dataset Setting.} 
In this setting, the model is trained on one dataset and evaluated on another dataset without fine-tuning. This is a more challenging setting than the \textit{zero-shot setting}, where the model not only deals with the \emph{unseen} classes, but also has to address the domain gap among different datasets.

\begin{figure*}[t]
  \centering
   \includegraphics[width=\linewidth]{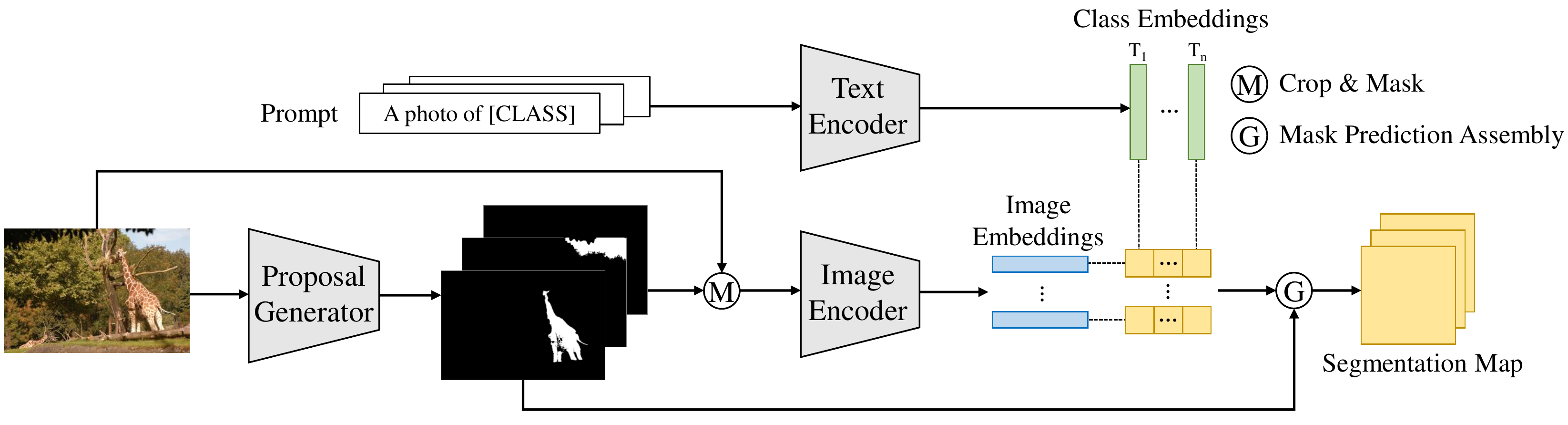}
   \caption{Overview of our two-stage open-vocabulary semantic segmentation framework. We reformulate and break down the open-vocabulary semantic segmentation into two steps: 1) training a mask proposal generator to generate a set of binary masks; 2) leveraging the pre-trained CLIP to classify each mask proposal. }
   \label{fig:proposal}
   \vspace{-1em}
\end{figure*}

\subsection{Revisiting CLIP}
CLIP~\cite{radford2021learning} is a powerful pre-trained vision-language model, which shows surprisingly strong performance in associating the visual and textual concepts. CLIP is a two-stream method: it contains an image encoder ${\cal E}_{\text{image}}$ and a text encoder ${\cal E}_{\text{text}}$. For any given image-text paired data $\{{\cal I}, {\cal T}\}$ , their semantic similarity can be estimated by computing the cosine distance between ${\cal E}_{\text{image}}({\cal I})$ and ${\cal E}_{\text{text}}({\cal T})$. 

The pre-trained CLIP model can be used to classify images by a given set of classes without fine-tuning, which is also known as zero-shot/open-vocabulary image classification. Specifically, the class names are injected into the pre-defined prompt template and fed into CLIP's text encoder to generate the class embeddings, e.g., a typical prompt template is `a photo of [CLASS]', where [CLASS] is replaced by the specific class name such as `person' and `cat'. The generated class embeddings are used as the classifier and the similarity with image embedding is computed for classification. 

In this work, we extend the compatibility of CLIP from \textit{image-level} zero-shot/open-vocabulary classification to \textit{pixel-level} open-vocabulary semantic segmentation, by exploring the use of a pre-trained CLIP model as a strong vision-category correspondent.

\section{Two-Stage Open-Vocabulary Semantic Segmentation}
Figure.~\ref{fig:proposal} shows an overview of our two-stage framework. Given an image, a set of mask proposals are first generated, and then each proposals is fed into an image encoder and compared with the class weights obtained by applying text encoder on the prompt class description to perform the classification. Finally, the mask prediction are assembled together to produce the final segmentation results. We will describe each component of our framework in the following.

\subsection{Mask Proposal Generation}
\label{sec:mask_proposal_generation}
We first introduce the mask proposal generation. In our work, we try three different methods to generate the mask proposals $\{{\cal M}^{p}_{k}\}$:

\noindent\textbf{GPB-UCM~\cite{arbelaez2010contour}}. 
This is a classical method to generate hierarchical segments by considering multiple low-level cues, e.g., brightness, color, texture, and local gradients. The generated segments of this approach are usually well aligned with the contour of objects.

\noindent\textbf{Selective Search~\cite{uijlings2013selective}}. 
This method can also generate hierarchical segments. Since this method can effectively localize objects, it is widely used in object detection systems~\cite{girshick2015fast,girshick2014rich}.

\noindent\textbf{MaskFormer~\cite{cheng2021per}}. 
This is a recently proposed method for supervised semantic segmentation. Unlike a fully convolution network that models the semantic segmentation as the pixel-wise classification problem, MaskFormer disentangles the semantic segmentation into two sub-tasks: predicting the segments at first and then classifying the category of each segment. We observe that the predicted segments by MaskFormer can be used as the mask proposals, and we empirically demonstrate (see~\cref{tab:proposal_gen}) that the MaskFormer trained on \emph{seen} classes can produce high-quality mask proposals on the \emph{unseen} classes. Therefore, we take this advantage of MaskFormer as our default mask proposal generator.

\subsection{Region Classification via CLIP}
\label{sec:region_classification}

\subsubsection{Two Strategies for Using CLIP.}
\label{sec:two_strategies}
There are two strategies to perform the region classification by utilizing the pre-trained CLIP:
\begin{itemize}
    \item The first strategy is to directly apply the CLIP image encoder on each mask proposals for classification. Specifically, given an image ${\cal I}$ and a mask proposal ${\cal M}^p$, the mask proposals are first binarized with a threshold of 0.5, and then apply the binarized ${\cal M}^p$ to image ${\cal I}$, erase the unused background and only crop foreground area. The masked image crop is resized to $224^2$ and then fed into CLIP for classification. However, since there is no extra training process, the training data of \emph{seen} classes cannot be utilized, resulting in inferior performance on \emph{seen} classes in the inference (see~\cref{tab:aba_classifier}).

    \item To utilize the training data of \emph{seen} classes, another approach is to retrain an image encoder. However, if we simply learn a set of new classifiers on the training data of \emph{seen} classes, the retrained image encoder has no generalization ability on \emph{unseen} classes since these classes have no corresponding classifiers. Therefore, we propose to use the features generated from the text encoder of the pre-trained CLIP model as the fixed classifier weights for the retrained image encoder. In this approach, the image encoder has a certain generalization ability to the \emph{unseen} classes since the image encoder is encouraged to embed the vision features into the same embedding space of the text encoder through the \emph{seen} classes.
    Notably, this approach can be easily integrated into the training process of the MaskFormer, by simply using the CLIP generated text features as the classifier weights of the MaskFormer, thus avoiding the need of training an additional image encoder.
\end{itemize}

The two strategies complement each other (see~\cref{tab:aba_classifier}), therefore we ensemble the results of these two strategies by default.
Given a mask proposal $\mathcal{M}^p$, we crop the foreground area $A_{fg}=\text{crop}(\mathcal{M}^p, \mathcal{I})$ (See Appendix for details), and compute its classification probability via CLIP vision encoder $E_{\text{vision}}$ and text encoder $E_{\text{text}}$:

\begin{equation}
    C_i(A_{fg}) = \frac{\text{exp}(\text{cosine}(E_{\text{vision}}(A_{fg})), E_{\text{text}}(\mathcal{C}_i)/\tau)}{\sum_{i}^{\text{\#class}} \text{exp}(\text{cosine}(E_{\text{vision}}(A_{fg}), E_{\text{text}}(\mathcal{C}_i))/\tau)}
\end{equation}

\noindent,where $\mathcal{C}_i$ is name of i-th class and temperature $\tau$=100. The classification probability of CLIP can be ensembled with supervised model trained on seen classes and then generate final mask results according to Sec.4.3.

\subsubsection{Prompt Design.}
\label{sec:text_promp}
The original CLIP is not designed for open-vocabulary semantic segmentation. How to design feasible text prompts need to be explored.  
\paragraph{Hand-Crafted Prompt.}
A simple approach is to re-use the hand-crafted prompts provided by CLIP which is originally designed for image classification on ImageNet-1K~\cite{deng2009imagenet}. There are 80 different prompts, each consisting of a natural sentence with a blank position for injecting the category names. Since these prompts are not originally designed for semantic segmentation, some of them may have a adverse effect. So we evaluate each of these prompts on training data to select one most helpful prompt for open-vocabulary semantic segmentation.

\paragraph{Learning-Based Prompt.}
Prompt learning~\cite{liu2021pre,zhou2021learning} recently showed great potential for adapting the pre-trained language/vision-language models on specific downstream tasks. We also explore this technique. Specifically, a prompt is a sequence of tokens. Each token belongs to one of the two types: $[P]$ indicates the prompt token and $[CLS]$ indicates the class token. A generalized prompt can be formulated as $[P]_0...[P]_m[CLS]$, where $m$ is the number of prompt token. In prompt learning, the prompt tokens $[P]_0...[P]_m$ are set as learnable parameters that can be trained on the \emph{seen} classes and generalized to the \emph{unseen} classes. 

\subsection{Mask Prediction Assembly}
\label{sec:mask_prediction_assembly}
Since the mask proposals may overlap each other, resulting in the possibility of some pixels being covered by several different mask proposals. Therefore, we employ a simple aggregation mechanism to generate semantic segmentation results from the mask predictions. Specifically, for a given pixel $q$, its predicted probability of being $i$-th category is defined as:

\begin{equation}
\small
    C_i(q) = \frac{{\cal M}_{k}^{p}(q) {C}_{k}^p(i)}{\sum{{\cal M}_{k}^{p}(q)}},
\end{equation}
where ${\cal M}_{k}^{p}(q)$ denotes the predicted probability of pixel $q$ in $k$-th mask proposal ${\cal M}_{k}^{p}$, and ${C}_{k}^p(i)$ is the predicted probability of mask proposals ${\cal M}_{k}^{p}$ belonging to $i$-th category. Note that the sum of $C_i(q)$ over all categories is not guaranteed to be 1, and pixel $q$ is classified to the category with highest predicted value.

\section{Fully Convolution Network Approach}
In addition to our proposed two-stage framework, a more conventional approach is to use the widely-used fully convolution network (FCN). As a dominant method in supervised semantic segmentation, FCN formulates the semantic segmentation as a pixel-wise classification problem. Specifically, given an image, FCN generates a high-resolution feature map, and a set of learned classifiers is applied on each pixel to produce segmentation predictions. Similar to our proposed two-stage framework, there are also two strategies to apply the CLIP on FCN framework:

\begin{itemize}
    \item Directly using the feature map generated by the CLIP vision encoder to perform pixel-wise classification. Note that in the original CLIP model, the feature of an image are represented by the feature of \texttt{[CLS]} token, not the feature map, and this difference may lead to performance degradation. In addition, the original CLIP model uses the image size of $224\times 224$ during pre-training, while semantic segmentation usually requires a higher image resolution (e.g., shorter size is $640$). Therefore, the direct use of high-resolution image during inference may lead to inferior performance due to inconsistency in image size. To alleviate this problem, we try to use the sliding window technique, which is widely used in previous works~\cite{chen2017deeplab} for performing multi-scale inference. We empirically found that it can improve performance and thus use it by default.
    
    \item The training data of the \emph{seen} classes cannot be utilized in the first strategy. Instead, we retrain an FCN-based vision encoder on \emph{seen} classes via the similar method introduced in Sec.~\ref{sec:two_strategies}. Specifically, we use the CLIP text encoder to generate a fixed classifier weight. Therefore, the retrained model can obtain a certain generalization ability to the \emph{unseen} classes.
\end{itemize}

As the same as the two-stage framework, we also ensemble the prediction of these two strategies by default if not specified.

\section{Experiments}
\subsection{Dataset and Evaluation Protocol}
\subsubsection{Dataset.}
We conduct extensive experiments on five challenging datasets to evaluate our method: COCO Stuff~\cite{caesar2018coco}, Pascal VOC 2012~\cite{everingham2011pascal}, Pascal Context~\cite{mottaghi2014role}, Cityscapes~\cite{cordts2016cityscapes}, and ADE20K~\cite{zhou2017scene}.

\noindent\textbf{COCO Stuff} is a large-scale dataset that contains 117k training images and 5k validation images. It contains annotations of 171 classes, 80 thing classes and 91 sutff classes respectively.

\noindent\textbf{Pascal VOC 2012} contains 11,185 training images and 1,449 validation images from 20 classes. The provided augmented annotations are used.

\noindent\textbf{Cityscapes} is a scene parsing dataset collected on urban streets, containing 5,000 finely annotated images and 20,000 coarsely annotated images. According to the common practices~\cite{cordts2016cityscapes}, we use 1,525 images of 19 classes in the finely annotated set for validation.

\noindent\textbf{Pascal Context} is an extensive dataset of Pascal VOC 2010, containing 4,998 training images and 5,005 validation images. We use the frequent 59 classes for validation.

\noindent\textbf{ADE20K} contains 20k training images, 2k validation images, and 3k testing images. There are two settings of 150 classes and 857 classes.

\subsubsection{Data Split.} 
For \emph{Cross-dataset setting}, we train our model on the COCO Stuff dataset and test on the validation set of the others. 
For \emph{Zero-shot setting}, we evaluate our method on COCO Stuff and Pascal VOC 2012. Following~\cite{xian2019semantic}, we divide the COCO Stuff dataset into 156 \emph{seen} classes and 15 \emph{unseen} classes and the Pascal VOC 2012 dataset into 15 \emph{seen} classes and 5 \emph{unseen} classes.

\subsubsection{Evaluation Protocol.} 
For \emph{cross-dataset setting}, we use the mean of class-wise intersection over union (mIoU) as major metric. For \emph{zero-shot setting}, we use harmonic mean IoU (hIoU) among the \emph{seen} classes and \emph{unseen} classes as major metric by following previous works~\cite{xian2019semantic,pastore2021closer} (see Appendix for detail definition). We also report the pixel-wise classification accuracy (pAcc) as a reference.

\subsection{Implementation Details} 
We conduct all experiments on 8$\times$Nvidia V100 GPUs. We train a MaskFormer~\cite{cheng2021per} model on the COCO Stuff dataset with ResNet-101 as the default backbone. An AdamW optimizer with the initial learning rate of 1e-4, weight decay of 1e-4 and a backbone multiplier of 0.1, and a poly learning rate policy with a power of 0.9 are used. The batch size is set to 32 for each GPU, and the total training iteration is 60K/120K for zero-shot setting and cross-dataset setting,respectively. If not specified, the MaskFormer model is only trained on \emph{seen} classes, and we use 100 mask proposals for both training and testing. For all other settings and hyper-parameters, we keep the original setting of MaskFormer without changes. CLIP with ViT-B/16 backbone is used by default if not specified. In text prompt tuning, the prompts are randomly initialized, and a SGD optimizer is used to train the learnable prompts. The learning rate is set to 0.02 and decayed according to the cosine learning rate policy, and the batch size is set to 32. We train 50 and 100 epochs for Pascal VOC and COCO Stuff, respectively. For Pascal VOC 2012 dataset, we use a batch size of 16 and a total training iteration of 20K, and keep all other setting as the same as the COCO Stuff dataset.

\begin{table}[t]
    \footnotesize
    \centering
    \caption{We train our model on COCO Stuff dataset and evaluate on other datasets (cross-dataset setting). The number in the parentheses after the dataset name represents class number. Both methods are tuned through the same prompt engineering~\cite{gu2021zero}.}
    \begin{tabular}{c|c|c|c|c}
    \toprule
     \diagbox{Method}{Dataset}& Cityscapes (19)& Pascal Context (59)&ADE20K (150) & ADE20K (847)\\
      \hline
      FCN &21.4&28.2 &14.9&4.1\\
      Ours &34.5 &47.7 &20.5&7.0\\
    \bottomrule
    \end{tabular}
    \label{tab:transfer_performance}
\end{table}

\begin{figtab}
\footnotesize
  \begin{minipage}[b]{0.48\linewidth}
    \centering
    \tabcaption{Comparison with other methods on COCO Stuff in the zero-shot setting.}
    \vspace{0.5em}
    \begin{tabular}{l|c|c|c}
    \toprule
    \multirow{2}{*}{Method}&\multirow{2}{*}{hIoU} &\multicolumn{2}{c}{mIoU}\\
      \cline{3-4}
     &&seen&unseen\\
    \hline
      SPNet~\cite{xian2019semantic}&16.8&20.5&14.3\\
      ZS3~\cite{bucher2019zero}&15.0&34.7&9.5\\
      CaGNet~\cite{gu2020context}&18.2&35.5&12.2\\
      FCN&20.9  &30.1&16.0\\
      Ours&37.8 &39.3&36.3\\
      \hline
      SPNet+ST~\cite{xian2019semantic}&30.3 &34.6&26.9\\
      ZS5~\cite{bucher2019zero}&16.2 &34.9&10.6\\
      CaGNet+ST~\cite{gu2020context}&19.5 &35.6&13.4\\
      STRICT~\cite{pastore2021closer}&32.6 &35.3&30.3\\
      Ours+ST&\textbf{41.5} &\textbf{39.6}&\textbf{43.6}\\
    \bottomrule
    \end{tabular}
    \label{tab:coco_stuff_res}
  \end{minipage}\quad
  \begin{minipage}[b]{0.48\linewidth}
    \centering
    \tabcaption{Comparison with other methods on Pascal VOC in the zero-shot setting.}
    \vspace{0.5em}
    \begin{tabular}{l|c|c|c}
    \toprule
    \multirow{2}{*}{Method}&\multirow{2}{*}{hIoU} &\multicolumn{2}{c}{mIoU}\\
      \cline{3-4}
     &&seen&unseen\\
    \hline
    SPNet~\cite{xian2019semantic}&25.1       & 73.3                                   & 15.0                 \\
    ZS3~\cite{bucher2019zero}&28.7        & 77.3                                    & 17.7                 \\
    CaGNet~\cite{gu2020context}&39.7      & 78.4                                              & 25.6                 \\
    FCN&50.7 & 85.5&36.0 \\
    Ours&77.5     & \textbf{83.5} & 72.5                \\
    \hline
    SPNet+ST~\cite{xian2019semantic}&38.8          & 77.80                                            & 25.8                 \\
    ZS5~\cite{bucher2019zero}&33.3              &            78                                    & 21.2                 \\
    CaGNet+ST~\cite{gu2020context}&43.7 & 78.6                                      & 30.3                 \\
    STRICT~\cite{pastore2021closer}&49.8  & 82.7                             & 35.6                 \\
    Ours+ST&\textbf{79.3}  & 79.2 & \textbf{78.1}                \\
    \bottomrule
    \end{tabular}
    \label{tab:pascal_voc_res}
  \end{minipage}
  \vspace{-1em}
\end{figtab}

\subsection{Comparison in Cross-Dataset Setting}
We first evaluate our method on the cross-dataset setting. The model is trained on the COCO Stuff dataset and then evaluated on other datasets without fine-tuning. ~\cref{tab:transfer_performance} clearly shows that our two-stage approach outperforms the FCN approach by a noticeable margin, demonstrating that our two-stage approach can better leverage the pre-trained CLIP model than the FCN approach. We do not list the result on Pascal VOC as its categories overlap much with the COCO Stuff dataset, and our method can achieve 88.4 mIoU.

\subsection{Comparison in Zero-Shot Setting}
We then compare our method with previous state-of-the-arts on Pascal VOC 2012 dataset and COCO Stuff dataset. Since some works reported the performance by applying the self-training techniques (denoted as ``ST"), we follow this practice and report the performance with or without self-training.

\begin{figure}[t]
  \centering
   \includegraphics[width=0.75\linewidth]{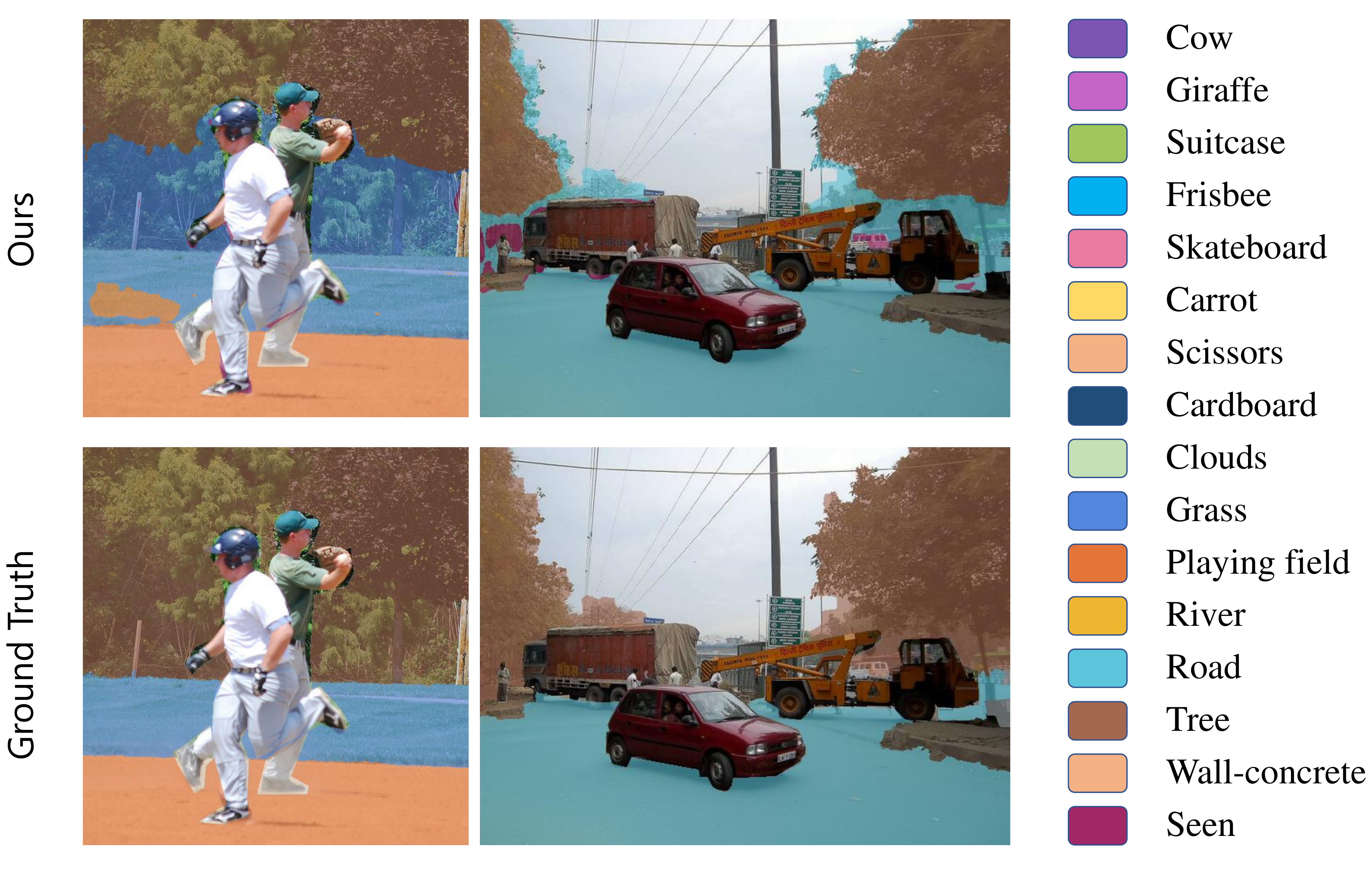}
   \caption{Qualitative results on COCO Stuff dataset. Only results of unseen classes are visualized. Predictions misclassified to seen classes are labeled with \textcolor[RGB]{161,40,100}{\emph{Seen}} color.}
   \label{fig:vis}
\end{figure}

\noindent\textbf{COCO Stuff.} 
\cref{tab:coco_stuff_res} shows the results. Compared with Pascal VOC 2012 dataset, COCO Stuff is more challenging. However, our approach still outperforms state-of-the-arts by a large margin. Specifically, without using the self-training, our method achieves 37.8 hIoU and 36.3 mIoU-unseen, outperforming the previous best method CaGNet~\cite{gu2020context} by +19.5 hIoU and +24.1 mIoU-unseen. By further employing the self-training, our method achieves 41.5 hIoU and 43.6 mIoU-unseen, outperforming the previous best method STRICT~\cite{pastore2021closer} by +8.9 hIoU and +13.3 mIoU-unseen. The qualitative results are shown in Figure.~\ref{fig:vis}. 

\noindent\textbf{Pascal VOC 2012.}
The results are shown in~\cref{tab:pascal_voc_res}. Without using the self-training, our method achieves 77.5 hIoU and 72.5 mIoU-unseen, outperforming the previous best method CaGNet~\cite{gu2020context} by a huge margin of +37.7 hIoU and +46.8 mIoU-unseen. By further employing the self-training, our method achieves 79.3 hIoU and 78.1 mIoU-unseen, outperforming the previous best method STRICT~\cite{pastore2021closer} by +29.5 hIoU and +42.5 mIoU-unseen.

\begin{table}[]
    \footnotesize
    \centering
    \caption{Study on how image encoder and pre-trained data affects the performance on Pascal VOC 2012 in the zero-shot setting.}
    \begin{tabular}{l|c|c|c|c|c}
    \toprule
    \multirow{2}{*}{Method} & Image & Pre-train & \multirow{2}{*}{hIoU} & \multicolumn{2}{c}{mIoU}\\
    \cline{5-6}
    & Encoder & Data & & seen & unseen \\
     \hline
     ZS3~\cite{bucher2019zero} & ResNet-101 & ImageNet & 28.7 & 77.3 & 17.7 \\
     CaGNet~\cite{gu2020context} & ResNet-101 & ImageNet & 39.7 & 78.4 & 25.6\\
     \hline
     \multirow{2}{*}{SPNet~\cite{xian2019semantic}} & ResNet-101 & ImageNet & 25.1 & 73.3 & 15.0\\
     & ResNet-101 & CLIP-VL & 33.4 & 74.1 & 21.5 \\
     \hline
     \multirow{3}{*}{Ours} & ResNet-101 &ImageNet & 49.5 & 71.1 & 38.0\\
     & ResNet-101 & CLIP-VL& 74.2 & \textbf{84.6} & 66.1 \\
     & VIT/B-16 & CLIP-VL& \textbf{77.5} & 83.5 & \textbf{72.5} \\
     \hline
    \end{tabular}
    \label{tab:fair_pascal}
\end{table}

\begin{figtab}
\footnotesize
  \begin{minipage}[t]{0.48\linewidth}
    \centering
    \tabcaption{Study of different proposal generation methods on COCO Stuff dataset.}
    \vspace{1em}
    \begin{tabular}{l|c|c|c}
    \toprule
      Method & hIoU & pAcc & mIoU-unseen\\
      \hline
      GPB-UCM~\cite{arbelaez2010contour}&10.9  &	9.5&	11.6  \\
      Sel. Search~\cite{uijlings2013selective}&11.0 &23.5&13.3\\
      MaskFormer~\cite{cheng2021per}&\textbf{28.2} &\textbf{48.4}&\textbf{29.7}\\
    \bottomrule
    \end{tabular}
    \label{tab:proposal_methods}
  \end{minipage}\quad
  \begin{minipage}[t]{0.48\linewidth}
    \centering
    \tabcaption{Evaluate the generalization ability of mask proposal generator.}
    \vspace{1em}
    \begin{tabular}{l|c|c|c}
    \toprule
      Training set& Test set & mIoU&pAcc \\
      \hline
      COCO Stuff&COCO Stuff &69.4&87.7\\
      ADE20K&COCO Stuff& 62.5&84.6\\
      \hline
      ADE20K&ADE20K	 &71.6&90.2\\
      COCO Stuff&ADE20K	 &64.4&87.7\\
    \bottomrule
    \end{tabular}
    \label{tab:proposal_gen}
    
  \end{minipage}
\end{figtab}

While our method outperforms other state-of-the-art zero-shot semantic segmentation methods, \textbf{how the larger pre-trained data and image encoder affects the performance is still unclear}. To study these impacts, we design a new implementation that enables our approach to only leverage ImageNet-1K classification data. Specifically, we train a vision-language model by only using ImageNet-1K: the class names of ImageNet-1K are treated as language inputs, and are encoded through a pure text encoder\footnote{We use SimCSE~\cite{gao2021simcse} as the text encoder trained on text data only.} to generate the classification weights. As shown in~\cref{tab:fair_pascal}, our method achieves 49.5 hIoU with ResNet-101 as backbone, which is much higher than other approaches. On the other hand, we also try to integrate the CLIP with SPNet, and we find it only achieves 33.4 hIoU, which is far from our method by using the same ResNet-101 backbone. Those experiments indicate that the surpassing performance of our method does not only come from larger pre-training data, but also our two-stage framework.

\subsection{Ablation Studies}
In this section, we validate the key designs of our method. If not specified, we report the performance on the COCO Stuff dataset with the MaskFormer model of ResNet-101 and CLIP of ViT-B/16 by using the \emph{zero-shot setting}.

\subsubsection{Different Mask Proposal Generation Methods.}
We evaluate the performance of the mask proposal generation methods by plugging them into our pipeline. To avoid the impact of the learnable classifier trained  on \emph{seen} classes, we perform the comparison by directly classifying the masked regions with the CLIP model. The results are shown in~\cref{tab:proposal_methods}, and the MaskFormer achieves better performance than the Selective Search and GPB-UCM. Note that even the other two methods are worse than the MaskFormer, they still achieve comparable performance compared with state-of-the-arts on mIoU-unseen.

\begin{figtab}
\footnotesize
  \begin{minipage}[t]{0.48\linewidth}
    \centering
    \tabcaption{Study of different region classification methods on COCO Stuff dataset.}
    \vspace{1em}
    \begin{tabular}{l|c|c|c}
    \toprule
      \multirow{2}{*}{Method} & \multirow{2}{*}{hIoU} & \multicolumn{2}{c}{mIoU}\\
      \cline{3-4}
      && seen & unseen\\
      \hline
      Retrained Vision Enc.  &8.7 &38.7&4.9  \\
      CLIP Vision Enc. &28.2 &26.8&29.7 \\
      Ensemble &\textbf{37.8}&\textbf{39.3}&\textbf{36.3} \\
    \bottomrule
    \end{tabular}
    \label{tab:aba_classifier}    
    
  \end{minipage}\quad
  \begin{minipage}[t]{0.48\linewidth}
    \centering
    \tabcaption{Evaluate the performance of different CLIP variants in our framework on COCO Stuff dataset with manual prompt.}
    \vspace{1em}
    \begin{tabular}{l|c|c}
    \toprule
      Backbone&hIoU  & mIoU-unseen\\
      \hline
      ResNet-50&15.2&16.0\\
      ResNet-101&13.8&12.5\\
      ViT-B/32&15.3&15.7\\
      ViT-B/16&\textbf{18.3} &\textbf{19.5}\\
    \bottomrule
    \end{tabular}
    \label{tab:clip_backbone}
  \end{minipage}
  \vspace{-2em}   
\end{figtab}

\begin{figtab}
 \footnotesize
  \begin{minipage}[t]{0.48\linewidth}
    \centering
    \tabcaption{Comparison with supervised method on COCO Stuff validation dataset. Sup: MaskFormer trained on both \emph{seen} classes and \emph{unseen} classes.}
    \vspace{1.0em}
    \begin{tabular}{l|c|c|c}
    \toprule
    Method& hIoU&mIoU & mIoU-unseen\\
      \hline        
      Sup&\textbf{49.4}&\textbf{42.6}&\textbf{62.6}\\
      Ours&37.8&39.2&36.3\\
      Ours+ST&41.5&39.9&43.6\\
    \bottomrule
    \end{tabular}
    \label{tab:coco_stuff_sup}
  \end{minipage}\quad
  \begin{minipage}[t]{0.48\linewidth}
    \centering
    \tabcaption{Comparison  with supervised method on the \emph{unseen} classes of COCO Stuff validation set. $\Delta$ is the difference in mIoU between things and stuff of \emph{unseen} classes.}
    \vspace{0.5em}
    \begin{tabular}{l|c|c|c|c}
    \toprule
    \multirow{2}{*}{Method}&\multicolumn{4}{c}{mIoU on \emph{unseen} classes}\\
    \cline{2-5}
    & all&thing&stuff&$\Delta$\\
      \hline
      Sup&\textbf{62.6}&\textbf{67.8}&\textbf{58.3}&9.5\\
      Ours&36.3&44.3&29.5&14.9\\
      Ours+ST&43.6&48.4&39.5&8.9\\
    \bottomrule
    \end{tabular}
    \label{tab:coco_stuff_sup_thing_stuff}
  \end{minipage}
  \vspace{-1em}
\end{figtab}

\subsubsection{Generalization of Mask Proposal Generator.} 
Although using MaskFormer to generate the mask proposals achieves excellent performance on zero-shot setting, it is still unknown whether Maskformer can produce good performance on the cross-dataset setting, i.e., training on one and testing on another dataset. Therefore, we evaluate the generalization ability of using MaskFormer to generate mask proposals between the COCO Stuff dataset and the ADE20K dataset.

In this experiment, we want to evaluate only the quality of the proposal without the effects of the region classifier. However, it is difficult to design a simple ``recall" metric for mask proposals in semantic segmentation similar to object detection. Because a segment can consist of multiple mask proposals, this may lead low recall while the final semantic segmentation result is still correct. Therefore, we designed an ``\textit{oracle}'' experiment to evaluate how these proposals affect the final performance of semantic segmentation. Specifically, for each mask proposal, its category is specified as the same as the ground-truth segment in which it has the largest overlap. In this case, the segmentation performance can fully reflect the proposal quality. 

The results are shown in~\cref{tab:proposal_gen}. We directly report the mIoU in this experiment because the \emph{seen} class cannot be defined between different datasets. We note that the MaskFormer model trained on COCO Stuff can produce good performance on ADE20K compared to the MaskFormer model directly trained on ADE20K with acceptable performance degradation, and vice versa. That demonstrates the generalization ability to use MaskFormer as the proposal generator.

\subsubsection{Different Strategies of Using CLIP.}
We study the two different strategies of using CLIP discussed in Sec.~\ref{sec:two_strategies}: retrained vision encoder or directly using CLIP vision encoder without tuning. The results are shown in~\cref{tab:aba_classifier}. The retrained vision encoder shows excellent performance on \emph{seen} classes, while its performance on \emph{unseen} classes is relatively low. In contrast, the CLIP vision encoder shows strong performance on \emph{unseen} classes while worse than retrained vision encoder on \emph{seen} classes by a large margin. By ensembling the two strategies, the performance on both \emph{seen} and \emph{unseen} classes is significantly improved, indicating the two strategies are complementary.

\subsubsection{Different CLIP Variants.} 
CLIP provides several variants with different network architectures and model sizes. We study how these models affect the performance of our method when using them as the region classifiers. We report the results without using the learnable prompt due to the high experimental overhead. The results are shown in~\cref{tab:clip_backbone}. We find that all models perform well and CLIP with ViT-B/16 achieves the best performance.

\subsubsection{Comparison with Supervised Baseline.}

We also compare our method with the supervised baseline on COCO Stuff. The supervised model is MaskFormer with ResNet-101 backbone which is trained on all classes, including \emph{seen} and \emph{unseen} classes. We report the results in~\cref{tab:coco_stuff_sup}. 
It is remarkable that while our method is worse than the supervised baseline by a large margin on mIoU-unseen and hIoU, the gap in mIoU is much close. That is because there are only 15 \emph{unseen} classes in the current dataset partition. For reference, there are 156 \emph{seen} classes.

To further explore the performance gap between our method and the supervised baseline, we split the \emph{unseen}  classes into \emph{things} and \emph{stuff}. The results are reported in~\cref{tab:coco_stuff_sup_thing_stuff}. We find the performance gap of our method between \emph{things} classes and the \emph{stuff} classes is significantly large than the supervised baseline, and self-training can significantly reduce the gap. This observation suggests that the classification ability of CLIP models is different in \emph{things} and \emph{stuff}, which may be due to the bias of the pre-trained dataset used by CLIP.

\section{Conclusion}
In this work, we propose a simple and effective two-stage framework for open-vocabulary semantic segmentation with the advanced pre-trained vision-language model. We reformulate and break down the open-vocabulary semantic segmentation into two steps: 1) training a mask proposal generator to generate a set of binary masks and 2) leveraging the pre-trained CLIP to classify each mask proposal. We conduct extensive experiments to verify our approach. Notably, the proposed framework outperforms previous state-of-the-arts of zero-shot semantic segmentation on Pascal VOC 2012 and COCO Stuff by large margins. Our work reveals the potential for using pre-trained vision-language models on open-vocabulary/zero-shot semantic segmentation and provides a strong baseline for this community to facilitate future research.

\appendix

\section{Definition of hIoU}
Following previous works~\cite{xian2019semantic,pastore2021closer}, harmonic mean IoU (hIoU) is defined among the \emph{seen} classes and \emph{unseen} classes as:

\begin{equation}
\small
     hIoU=\frac{2*{mIoU}_{seen}*{mIoU}_{unseen}}{{mIoU}_{seen}+{mIoU}_{unseen}}.
\end{equation}

\section{Sliding Window Testing in Fully Convolutional Network}
We study the different inference methods in this section for Fully Convolutional Network(FCN). For a fair comparison, we use ResNet-101 in FCN and ViT-B/16 in CLIP, same as our two-stage framework. Table.~\ref{tab:fcn} shows the results. The FCN without sliding window test achieved 11.7 hIoU and 10.4 mIoU-unseen. In comparison, employing the window test improved the performance by +9.2 on hIoU and +5.6 on mIoU-unseen. This significant difference in performance is caused by the inconsistent image size between pre-training and testing of the CLIP model. Although the sliding window test can strengthen the FCN approach, it is still worse than our two-stage approach by -16.8 on hIoU and -20.3 on mIoU unseen, indicating that our two-stage framework is more suitable for the CLIP model.

\begin{table}[]
    \footnotesize
    \centering
    \caption{Performance of FCN approach on COCO Stuff dataset under the \emph{zero-shot} setting. SW: Sliding Window Testing, each image is splited into several $224\times 224$ patches.}
    \begin{tabular}{c|c|c|c}
    \toprule
    Method & hIoU & pACC & mIoU-unseen\\
  \hline
    FCN~\cite{long2015fully} &11.7& 54.9 & 10.4\\
    FCN + SW~\cite{long2015fully} &20.9 & 50.8 &16.0  \\
    \hline
    Ours & \textbf{37.7}&\textbf{60.3}&\textbf{36.3}\\
    \bottomrule
    \end{tabular}
    \label{tab:fcn}
    \vspace{-2em}
\end{table}

\section{Prompt Engineering for Image and Text}
\subsection{Prompt Engineering for Image}

\begin{figure}[t]
  \centering
  \includegraphics[width=0.8\linewidth]{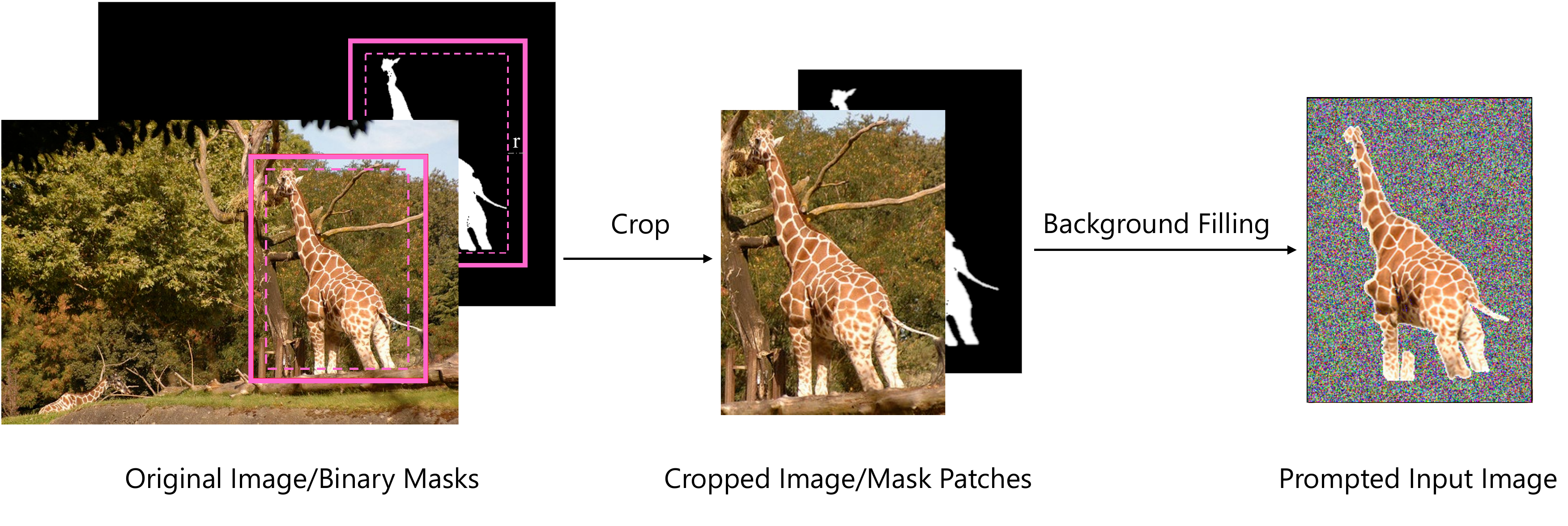}
  \caption{Pipeline of image prompt engineering. The pink dotted box is the minimum bounding box of the foreground region. The pink solid box is the expanded bounding box of the dotted box by the ratio $r$. The prompted image is the input of the CLIP image encoder.}
   \label{fig:pipe}
\end{figure}
\begin{figure}[t]
  \centering
  \includegraphics[width=0.8\linewidth]{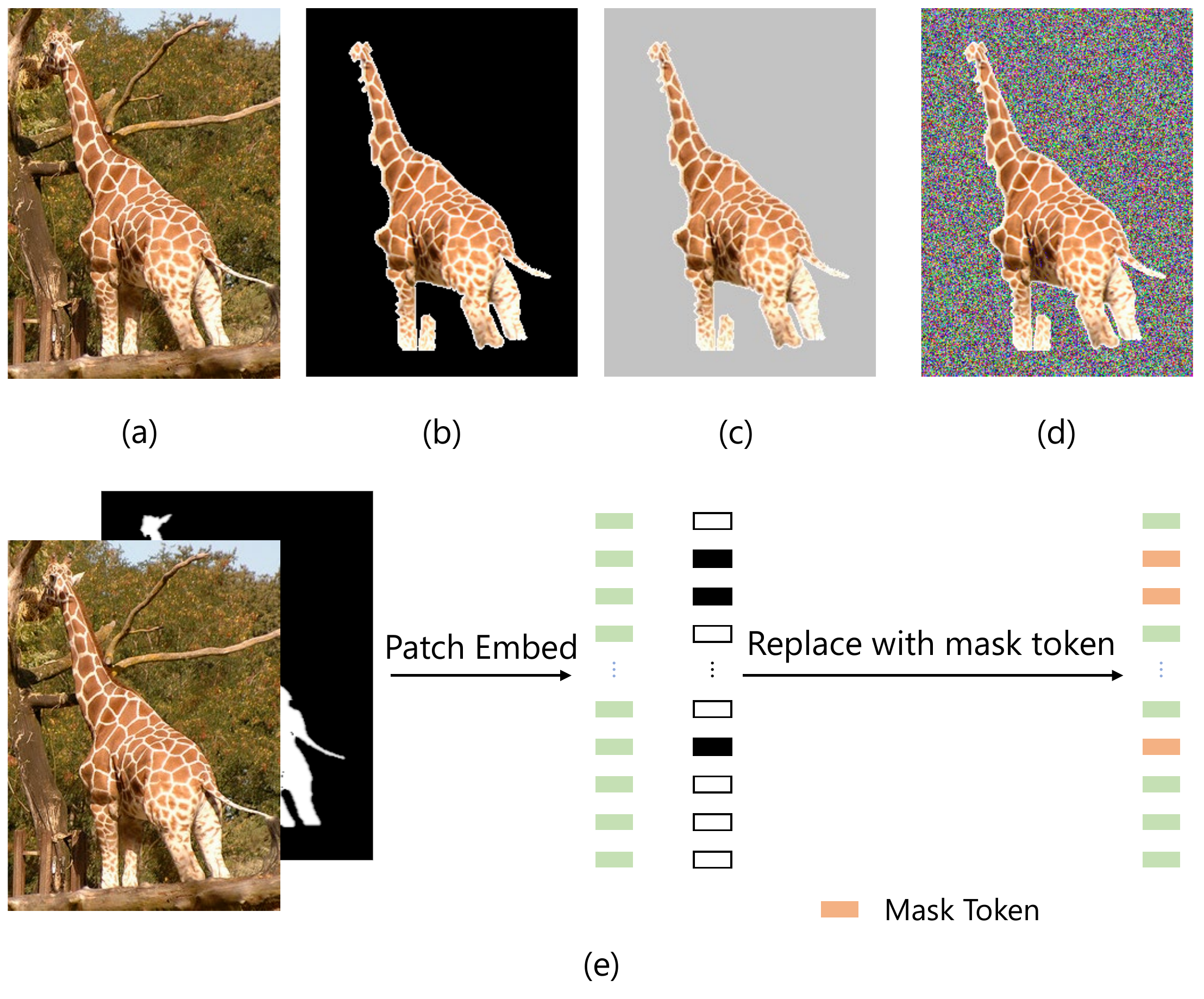}
  \caption{Choices for background filling. a) Preserving the context pixels; b) Filling the background pixels with zero; c) Filling the background pixels with the mean values of the dataset~\cite{radford2021learning}; d) Filling the background pixels with learnable pixel prompts. The prompts are tuned on \emph{seen} classes; e) Filling the background patches with mask token. The mask token is tuned on \emph{seen} classes.}
   \label{fig:bgfill}
\end{figure}

As the CLIP model is trained with low-resolution realistic images, given a mask proposal $\mathcal{M}^p$, and the input image $I$, it is a problem how to extract the visual representation of the proposal with the CLIP model through the proper way, which we call image prompt engineering. The whole process is shown in Figure.~\ref{fig:pipe}. We crop the image with the bounding boxes of $\mathcal{M}^p$ and expand the bounding boxes by a ratio $r$ to involve more context information. And then, we fill the background pixels with $0$ values in the proposal with some patterns. We studied four choices for such patterns: a) Keep the background pixels unchanged; b) Fill the background pixels with manually designed values; c) Fill the background pixels with learnable values; d) Fill the background patches with mask token, presented in Figure.~\ref{fig:bgfill}. The results are shown in~\cref{tab:bg_fill_result}. The value filled in the background area can greatly affect the segmentation performance. 
Though our exploration to learn proper image prompts failed to achieve improvement like text~\cite{liu2021pre,zhou2021learning}, it is still an interesting problem for future research.

\begin{table}[]
    \footnotesize
    \centering
    \caption{\footnotesize Study the effects of background filling. Filling background with learnable pixel prompts or mask token is very unstable and usually leads to negative impact. }
    \begin{tabular}{l|c|c|c}
    \toprule
      \multirow{2}{*}{Prompt} & \multirow{2}{*}{hIoU} & \multicolumn{2}{c}{mIoU}\\
      \cline{3-4}
      && seen & unseen\\
      \hline
     Preserving & 9.3&8.9 & 9.5  \\
     Zero & 17.2&16.3&18.2 \\
     Mean Values &18.3&17.3&19.5 \\
     Pixel Prompts &Failed&-&- \\
     Mask Token&Failed&-&- \\ 
     \bottomrule
    \end{tabular}
    \label{tab:bg_fill_result}
    \vspace{-2em}
\end{table}

\subsection{Prompt Engineering for Text}

\begin{figtab}
\footnotesize
\begin{minipage}[t]{0.48\linewidth}
    \centering
    \tabcaption{Study the effect of prompt length and sample number of each category for prompt learning.}
    \vspace{0.5em}
    \begin{tabular}{c|c|c}
    \toprule
      Prompt Len& \#Sample & Unseen Acc\\
      \hline
      \multirow{4}{*}{16}&16&28.9  \\
      &32&30.0\\
      &64&29.5\\
      &all&25.5\\
      \hline
      \multirow{4}{*}{32}&16&32.1\\
      &32&\textbf{32.8}\\
      &64&31.0\\
      &all&27.6\\
    \bottomrule
    \end{tabular}
    \label{tab:promt_hyper}
\end{minipage}\quad
\begin{minipage}[t]{0.48\linewidth}
    \centering
    \tabcaption{Manually designed prompt v.s. Learnable prompt.}
    \vspace{1em}
    \begin{tabular}{l|c|c|c}
    \toprule
      \multirow{2}{*}{Prompt} & \multirow{2}{*}{hIoU} & \multicolumn{2}{c}{mIoU}\\
      \cline{3-4}
      && seen & unseen\\
      \hline
      Manual&18.3&17.3&19.5\\
      Learnable&\textbf{28.2} &\textbf{26.8}&\textbf{29.7}\\
    \bottomrule
    \end{tabular}
    \label{tab:ab_learn_prompt}
\end{minipage}
\end{figtab}

We compare two prompt tuning methods described in Sec.~4.2. The results are shown in~\cref{tab:ab_learn_prompt}. The learnable prompt outperforms the manually searched prompt by +9.9 hIoU, clearly showing the power of the learnable prompt. In addition, although the learnable prompt is only trained on \emph{seen} classes, we notice that it achieves similar improvement on \emph{seen} classes and \emph{unseen} classes (+9.6 mIoU-seen and +10.2 mIoU-unseen), indicating the learnable prompt has a strong generalization ability to the \emph{unseen} class. 

We further study how prompt length and training data size affect the performance of learnable prompts by training on \emph{seen} classes and testing on \emph{unseen} classes. ~\cref{tab:promt_hyper} shows that using 32 samples for each category reaches the best performance, in either prompt length of 16 or 32, and more training samples will degrade the performance. We speculate that more samples may lead to the over-fitting issue, which is also reported by other prompt learning attempts~\cite{zhou2021learning}.

\begin{table}[]
    \footnotesize
    \centering
    \caption{\footnotesize Study the effects of MaskFormer and CLIP.}
    \begin{tabular}{l|c|c|c|c|c}
    \toprule
    \multirow{2}{*}{Method} & Image & Pre-train & \multirow{2}{*}{hIoU} & \multicolumn{2}{c}{mIoU}\\
    \cline{5-6}
    & Encoder & Data & & Seen & Unseen \\
     \hline
     SPNet[45] & \multirow{1}{*}{R-101} & \multirow{1}{*}{ImageNet} & \multirow{1}{*}{25.1} & \multirow{1}{*}{73.3} & \multirow{1}{*}{15.0}\\
     \hline
     \multirow{1}{*}{FCN} &  VIT/B-16 & CLIP-VL & 50.7 & \textbf{85.5} & 36.0 \\
     \hline
     \multirow{3}{*}{MaskFormer} & \multirow{1}{*}{R-101} & \multirow{1}{*}{ImageNet} & \multirow{1}{*}{49.5} & \multirow{1}{*}{71.1} & \multirow{1}{*}{38.0}\\
      & R-101 & CLIP-VL& 74.2 & 84.6 & 66.1 \\
      & VIT/B-16 & CLIP-VL & \textbf{77.5} & 83.5 & \textbf{72.5} \\
     \bottomrule
    \end{tabular}
    \label{tab:fair_pascal}
    \vspace{-1em}
\end{table}

\section{Detailed Study on MaskFormer and CLIP} 
The ablations have been studied in Table.~3 and Table.~4. We re-organize the results as shown in Table.~\ref{tab:fair_pascal}. We can conclude: 1) MaskFormer outperforms FCN by +26.8 hIoU (2-th row vs 5-th row); 2) CLIP pre-training outperforms ImageNet pre-training by +24.7 hIoU (3-rd row vs 4-th row); 3) Our method outperforms SPNet by +24.4 hIoU with the same pre-training data (1-st row vs 3-rd row).

\begin{table}[]
    \footnotesize
    \centering
    \caption{\footnotesize Results of different seen/unseen splits on COCO Stuff. 0* denotes the split used in previous work~\cite{xian2019semantic}.}
    \begin{tabular}{l|c|c|c|c|c|c|c}
    \toprule
    \multirow{2}{*}{Split}&\multicolumn{2}{c|}{Thing/Stuff Ratio}&\multirow{2}{*}{hIoU}&\multicolumn{4}{c}{mIoU$_{\text{unseen}}$}\\
    \cline{2-3} \cline{5-8}
    &Seen&Unseen&&All&Thing & Stuff & $\Delta$ \\
     \hline
     0*&0.88&0.87&37.8&36.3 &44.3 &29.5 &14.9\\
     \hline
     1&0.95&0.36&31.9&25.6&44.8&18.6&26.2\\
     2&0.93&0.50&30.9&	24.3&	41.5&	15.6&25.9\\
     3&0.86&1.14&36.6&32.9&38.8&26.2&12.6\\
     \bottomrule
    \end{tabular}
    \label{tab:randomness_of_split}
    \vspace{-2em}
\end{table}

\section{The Randomness of the Data split}
In the experiments under the zero-shot setting, we use the official unseen/seen split as~\cite{xian2019semantic} for a fair comparison. The thing/stuff ratio is 0.88 for seen and 0.87 for unseen classes. To study the impacts of different splits, we conduct studies on randomly generated seen/unseen splits in Table.~\ref{tab:randomness_of_split}. We find a more balanced unseen thing/stuff ratio yields higher hIoU.

\section{Visualization of Results under Cross-dataset Setting}
\begin{figure}[t]
  \centering
     \includegraphics[width=0.95\textwidth]{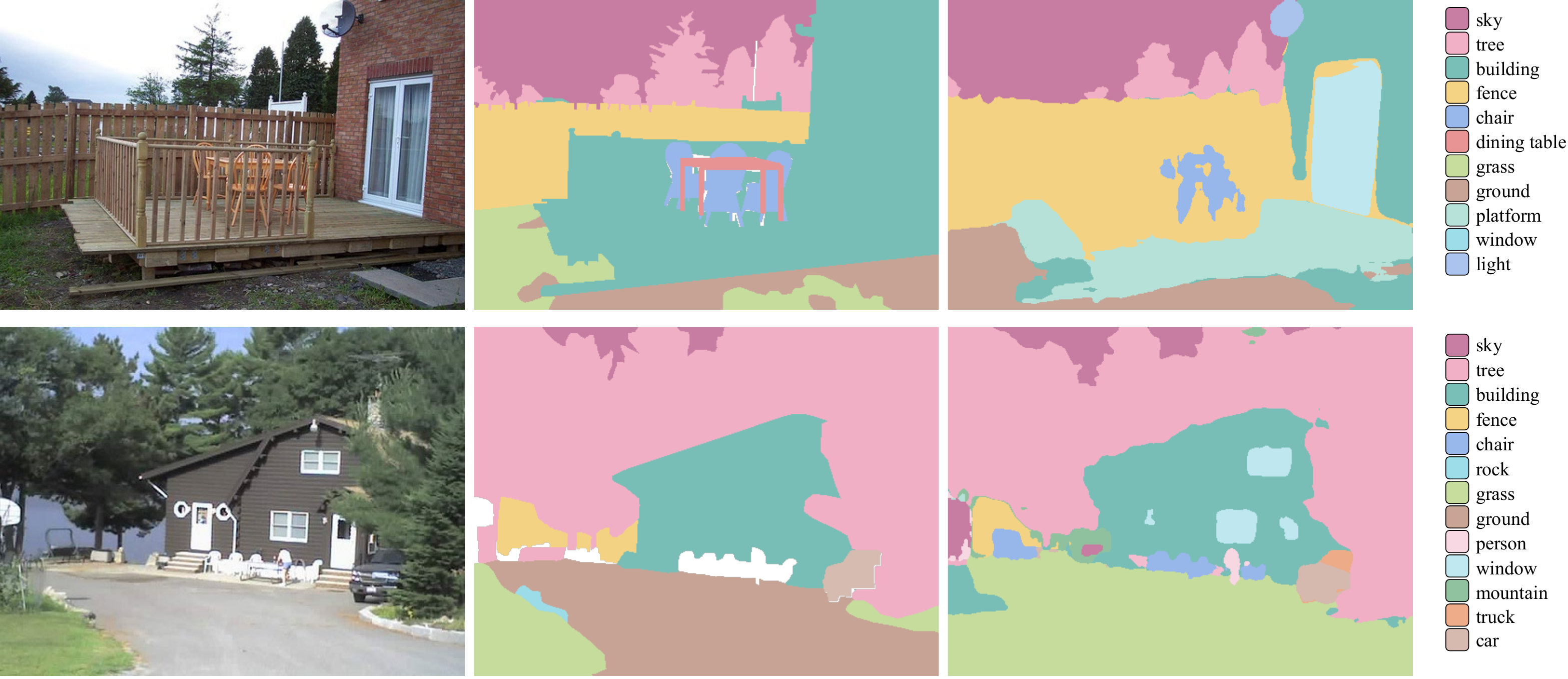}
     \caption{Qualitative results on Pascal Context dataset under \emph{cross-dataset} setting. From left to right are the original input images, the ground truth semantic segmentation maps and the predictions. The white areas in the ground truth maps is ignored during annotating. }
     \label{fig:supp_pascal}
     \vspace{-2em}
\end{figure}

\begin{figure}[t]
    \centering
    \includegraphics[width=0.95\linewidth]{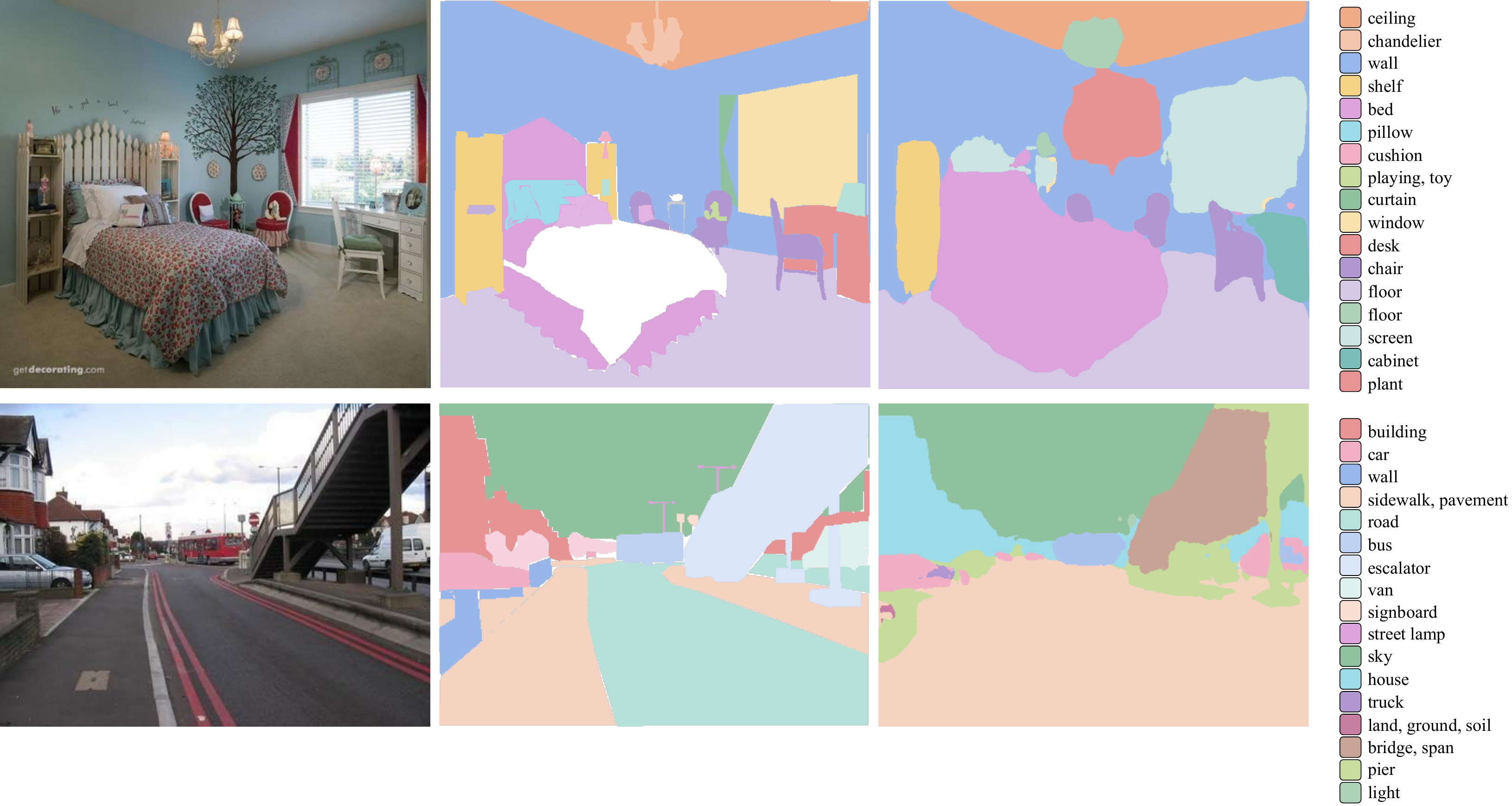} 
     \caption{Qualitative results on ADE20k dataset under \emph{cross-dataset} setting. The original input images, the ground truth semantic segmentation maps, and the predictions are left to right. The white areas in the ground truth maps is ignored during annotating. }
     \label{fig:supp_ade}
\end{figure}

We illustrate more qualitative results in Figure.~\ref{fig:supp_pascal},~\ref{fig:supp_ade} under the cross-dataset setting.


\clearpage
%
%
\bibliographystyle{splncs04}
\bibliography{reference.bib}
\end{document}